%% file: main.tex
\documentclass[journal]{IEEEtran}
\usepackage{cite}
\usepackage{amsmath,amssymb,amsfonts}
\usepackage{amsthm}
\usepackage{algorithm}
\usepackage{algorithmic}
\usepackage{graphicx}
\usepackage{textcomp}
\usepackage{color}
\usepackage{xspace}
\usepackage{booktabs}
\usepackage{longtable}
\usepackage{multirow}
\usepackage{hyperref}
\hypersetup{colorlinks=true,
            linkcolor=red,
            anchorcolor=blue,
            citecolor=blue}
\usepackage[T1]{fontenc}
\usepackage{float}
\usepackage{latexsym}
\usepackage{soul}
\usepackage{makecell}
\usepackage{times}
\usepackage{epsfig}
\usepackage{gensymb}
\usepackage[table]{xcolor}
\usepackage{epstopdf}
\usepackage{diagbox}
\usepackage{bm}

\input{math_definition.tex}

\newlength\savewidth\newcommand\shline{\noalign{\global\savewidth\arrayrulewidth
  \global\arrayrulewidth 1pt}\hline\noalign{\global\arrayrulewidth\savewidth}}

\newcommand{\methodname}{\textsc{{Phasor}}\xspace}
\newcommand{\modelname}{{AR-MoE}\xspace}
\newcommand{\alignname}{{IP-REPA}\xspace}

\renewcommand{\paragraph}[1]{\noindent\textbf{#1}\hspace{2mm}}

\newcommand{\papertitle}{\methodname: Anatomy- and Phase-Consistent Volumetric Diffusion for CT Virtual Contrast Enhancement}

\def\BibTeX{{\rm B\kern-.05em{\sc i\kern-.025em b}\kern-.08em
    T\kern-.1667em\lower.7ex\hbox{E}\kern-.125emX}}

\begin{document}

\title{\papertitle}

\author{
Zilong~Li,
Dongyang~Li,
Chenglong~Ma,
Zhan~Feng,
Dakai~Jin,
Junping~Zhang,~\IEEEmembership{Senior~Member,~IEEE},
Hao~Luo,
Fan~Wang,
Hongming~Shan,~\IEEEmembership{Senior~Member,~IEEE},
\thanks{Z. Li and J. Zhang are with the Shanghai Key Lab of Intelligent Information Processing, School of Computer Science and Artificial Intelligence, Fudan University, Shanghai 200433, China. (e-mail: longzilipro@gmail.com; jpzhang@fudan.edu.cn)}
\thanks{C. Ma and H. Shan are with the Institute of Science and Technology for Brain-inspired Intelligence and MOE Frontiers Center for Brain Science, Fudan University, Shanghai 200433, China, and also with Shanghai Center for Brain Science and Brain-inspired Technology, Shanghai 201602, China. (e-mail: clma24@m.fudan.edu.cn; hmshan@ieee.org)}
\thanks{Z. Feng is with the Department of Radiology, First Affiliated Hospital of College of Medical Science, Zhejiang University, Hangzhou, China. (e-mails: gerxyuan@zju.edu.cn)}
\thanks{Z. Li, D. Li, D. Jin, H. Luo, and F. Wang are with Alibaba DAMO Academy, China. (e-mails: longzili.lzl@alibaba-inc.com; yingtian.ldy@alibaba-inc.com; dakai.jin@alibaba-inc.com; michuan.lh@alibaba-inc.com; fan.w@alibaba-inc.com)}
% \thanks{Corresponding authors: Hao Luo (michuan.lh@alibaba-inc.com) and Hongming Shan (hmshan@fudan.edu.cn).}
}

\maketitle

\input{secs/abs.tex}

\input{secs/intro.tex}

\input{secs/relate.tex}

\input{secs/method.tex}

\input{secs/result.tex}

\input{secs/discuss_conclu.tex}


\end{document}

%% file: math_definition.tex
\usepackage{amsmath,amsfonts}
\usepackage{amsthm}
\usepackage{amssymb,amsopn}
\usepackage{bm} % bold symbol

\newcommand{\mat}[1]{\mathbf{#1}} % matrix

\newcommand{\eat}[1]{}

\newcommand{\ie}{\emph{i.e.}\xspace}
\newcommand{\eg}{\emph{e.g.}\xspace}

%% file: secs/abs.tex
\begin{abstract}

Contrast-enhanced computed tomography (CECT) is pivotal for highlighting tissue perfusion and vascularity, yet its clinical ubiquity is impeded by the invasive nature of contrast agents and radiation risks.
While virtual contrast enhancement (VCE) offers an alternative to synthesizing CECT from non-contrast CT (NCCT), existing methods struggle with \emph{anatomical heterogeneity} and \emph{spatial misalignment}, leading to inconsistent enhancement patterns and incorrect details.
This paper introduces \methodname, a volumetric diffusion framework for high-fidelity CT VCE. By treating CT volumes as coherent sequences, we leverage a video diffusion model to enhance structural coherence and volumetric accuracy.
To ensure anatomy-phase consistent synthesis, we introduce two complementary modules. 
First, anatomy-routed mixture-of-experts (\modelname) anchors distinct enhancement patterns to anatomical semantics, with organ-specific memory to capture salient details.
Second, intensity-phase aware representation alignment (\alignname) highlights intricate contrast signals while mitigating the impact of imperfect spatial alignment.
Extensive experiments across three datasets demonstrate that \methodname{} significantly outperforms state-of-the-art methods in both synthesis quality and enhancement accuracy.
Code will be publicly available upon acceptance.

\end{abstract}

%% file: secs/intro.tex
\section{Introduction}

Contrast-enhanced computed tomography (CECT) stands as a cornerstone of radiological diagnosis, revealing critical hemodynamic patterns as well as organ and lesion details often imperceptible in non-contrast CT (NCCT)~\cite{ct,damo_noncontrast,damo_panda}. However, its widespread application in screening and emergency settings is restricted by the invasive nature of contrast agents, the risk of allergic reactions, and cumulative radiation exposure.

Consequently, virtual contrast enhancement (VCE) has emerged as a compelling paradigm for synthesizing high-fidelity CECT from NCCT~\cite{cect_multiphase_relation,zhong2025ckap}. Early works leveraged GANs~\cite{lyu2023cta, zhong2025ckap} or diffusion models~\cite{ozbey2023unsupervised, meng2024multi}, and incorporated phase embeddings~\cite{cect_multiphase_relation, cect_miccai_shuoli} and structure-aware optimization~\cite{maformer, zongwei_ce2ct} to improve synthesis fidelity.

Despite these advances, robust high-fidelity VCE remains challenging. 
First, most existing methods formulate VCE as slice-wise image-to-image translation, which overlooks long-range volumetric dependencies and often yields inconsistent 3D structures.
Second, even with volumetric modeling, achieving anatomy-phase consistent synthesis is hindered by two core challenges: 
\textbf{(i)} \emph{anatomical heterogeneity}, where organs such as the liver and kidney exhibit distinct enhancement dynamics. This imposes a heavy burden on the model to implicitly localize anatomical structures to disentangle these patterns, leading to inaccurate enhancement; 
and 
\textbf{(ii)} \emph{spatial misalignment}, 
where physiological motions induce inevitable misalignment at boundaries such as bone or air, yielding intensity discrepancies exceeding hundreds of Hounsfield Units (HU). This large spatial error overshadows subtle enhancement signals of only 20$\sim$100 HU, corrupting the phase-specific contrast details.

\begin{figure*}[t]
    \centering    
    \includegraphics[width=0.95\linewidth]{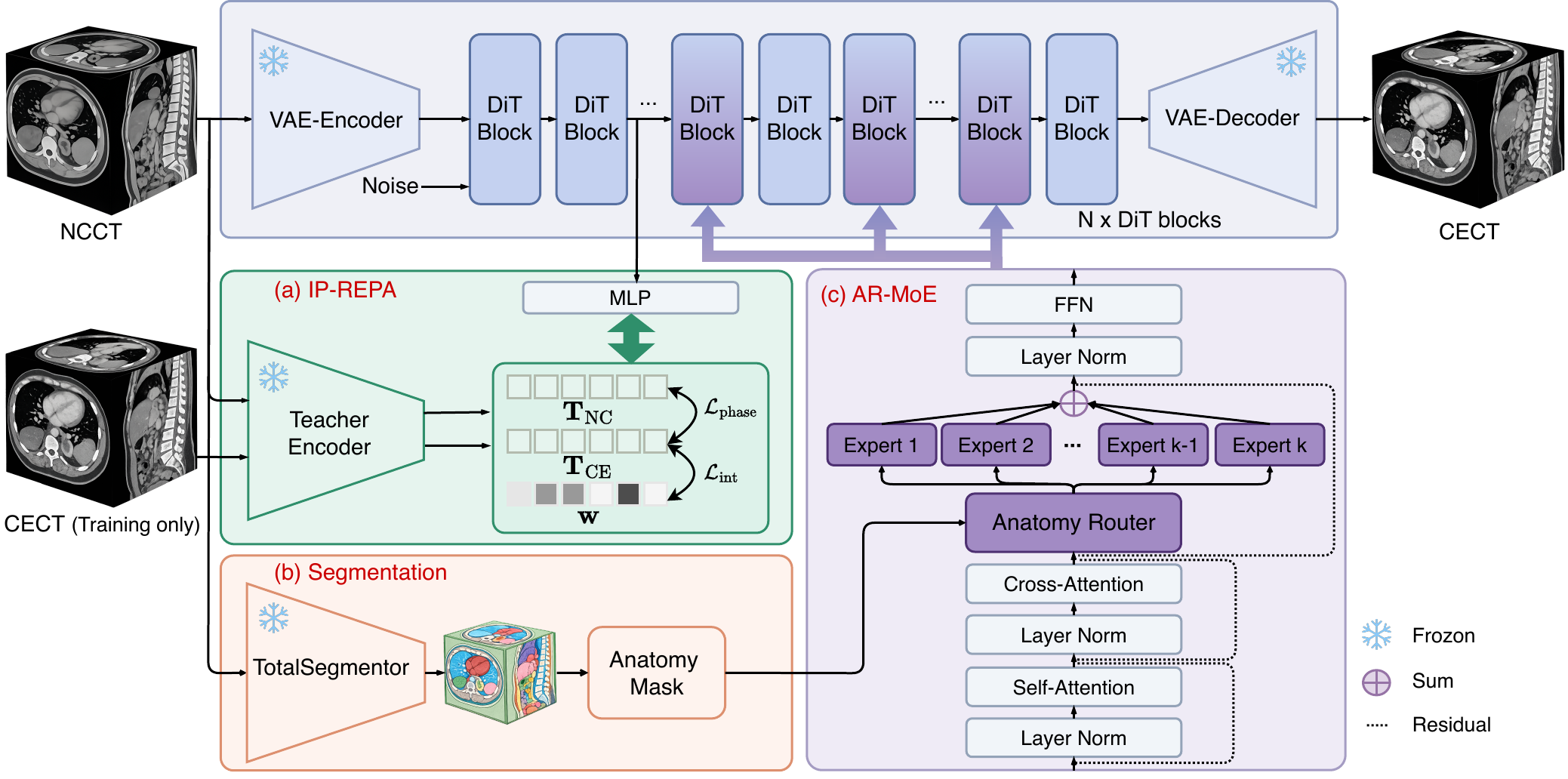}
    \caption{Overview of \methodname{}, which takes NCCT as input and synthesizes CECT. (a) \alignname{} aligns DiT features with representations extracted from a teacher encoder during training. (b) TotalSegmentor provides organ-wise anatomy masks. (c) \modelname{} routes tokens to specialized experts based on the anatomy masks.}
    \label{fig:phasor_framework}
    \vspace{-6pt}
    % \vspace{4pt}
\end{figure*}

In this work, we introduce \methodname{}, a volumetric diffusion framework designed for anatomy-phase consistent CT VCE. 
% Overall Framework
First, we leverage a video diffusion transformer (DiT) by treating CT volumes as coherent sequences. This design not only enforces global structural coherence across slices but also significantly accelerates inference throughput compared to slice-wise approaches~\cite{wan21, ctgen_trace}.
% Method 1: SR-MoE
Second, to ensure anatomy-consistent enhancement, we propose an anatomy-routed mixture-of-experts (\modelname). 
\modelname{} dynamically routes tokens to organ-specific experts, thus anchoring distinct enhancement patterns to anatomical semantics.   
Each expert is further augmented with organ-specific memory to capture salient details from the NCCT condition.
% Method 2: Alignment
Third, to ensure phase-consistent enhancement, we introduce an intensity-phase aware representation alignment (\alignname). 
This strategy highlights intricate phase-related contrast signals while mitigating the impact of spatial misalignment, guiding the model to focus on phase-specific enhancement rather than registration noise.
Experimental results across three datasets demonstrate that \methodname{} significantly outperforms state-of-the-art methods in both synthesis quality and enhancement accuracy. Notably, our framework enables robust training on imperfectly registered data and supports simultaneous multi-phase synthesis.

The main contributions of this work are summarized as follows.  
1) We propose \methodname{}, a novel volumetric diffusion framework that leverages a video diffusion model to enhance global structural coherence and high-fidelity volumetric synthesis.  
2) We design \modelname{} to ensure anatomy-consistent enhancement, anchoring distinct enhancement patterns to anatomical semantics.  
3) We introduce \alignname{} to ensure phase-consistent enhancement, highlighting fine-grained contrast details while mitigating the impact of spatial misalignment.  
4) Extensive experimental results demonstrate that \methodname{} achieves state-of-the-art performance across synthesis quality and enhancement accuracy.

%% file: secs/relate.tex
\section{Related Work}

\subsection{Virtual Contrast Enhancement}
Virtual Contrast Enhancement (VCE) aims to synthesize contrast-enhanced images directly from non-contrast acquisitions, reducing radiation exposure and enabling large-scale screening~\cite{dinggang_mrigen,dinggang_petgenct}.
Early approaches largely relied on GAN-based translation for various cross-modality tasks, such as NCCT-to-CECT, cross-sequence MRI, and MRI-to-PET synthesis~\cite{lyu2023cta,maformer,dinggang_unisyn,qiangzhang_mrigen}.  
However, GANs inherently suffer from mode collapse and limited image quality. Given the high dynamic range and high resolution of CT images, these models frequently struggle to synthesize delicate textures required for clinical diagnosis. 
Recently, diffusion models have gained prominence as an alternative to GANs, delivering markedly improved synthesis quality~\cite{zhu2023make,wang2025meddiffusion,ozbey2023unsupervised}. 
Nevertheless, most existing VCE methods continue to treat the problem as a generic slice-wise translation~\cite{zongwei_ce2ct}, while also neglecting inter-slice relationships and complex organ-wise enhancement dynamics. 
To mitigate this, recent works introduce structural priors, utilizing techniques such as self-supervised learning~\cite{maformer}, spatially adaptive normalization~\cite{aldm}, and segmentation-based losses~\cite{zongwei_ce2ct}. 
Despite this progress, achieving high-fidelity CT VCE that simultaneously ensures structural coherence and accurate contrast enhancement remains unsolved. 
This paper introduces \methodname{}, a novel framework that \emph{treats CT slices as video frames}, and \emph{tackles anatomical heterogeneity and enhances fine-grained contrast details, simultaneously}.

\subsection{Volumetric Latent Diffusion Models}
Latent Diffusion Models (LDMs)~\cite{ldm} have been extended to volumetric domains~\cite{blattmann2023align, wan21}, where the CT $z$-axis functions as a pseudo-temporal sequence~\cite{ctgen_text2ct,ctgen_trace}.
This paradigm facilitates holistic synthesis, effectively preserving long-range anatomical dependencies (\eg, continuous vessels) often severed by slice-wise approaches.
However, generic video backbones—prioritizing stylistic continuity over rigorous structural precision—frequently struggle with the fine-grained fidelity required for medical tasks~\cite{zongwei_ce2ct}.
Recently, representation alignment strategies like RePA~\cite{repa} and VideoRePA~\cite{videorepa} have emerged to improve semantics and physical consistency. 
Inspired by these advances, we \emph{introduce video diffusion transformers for CT VCE} and \emph{design \alignname{}} to enhance fine-grained contrast details.

%% file: secs/method.tex
\section{Method}

\subsection{Overview of \methodname}

As shown in Fig.~\ref{fig:phasor_framework}, \methodname{} aims to synthesize a high-fidelity CECT volume $\mat{X}_{\text{CE}} \in \mathbb{R}^{H\times W\times D}$ from an input NCCT volume of the same size $\mat{X}_{\text{NC}}$, where $H$, $W$, and $D$ denote the height, width, and depth (\ie, number of slices) of the volume, respectively. 
Our framework is built upon WAN 2.1~\cite{wan21}, including a 3D VAE fine-tuned on CT data for volumetric compression and a Diffusion Transformer (DiT) backbone for generation. 
Specifically, the 3D VAE encoder $\mathcal{E}$ compresses input volumes into latent representations, which are then patchified and projected into a unified token space, yielding $\mat{Z}_{\text{NC}}$ and $\mat{Z}_{\text{CE}} \in \mathbb{R}^{N \times C}$ as the latent embedding of $\mat{X}_{\text{NC}}$ and $\mat{X}_{\text{CE}}$, where $N$ is the sequence length and $C$ is the embedding dimension.
The DiT backbone drives the generation via Rectified Flow~\cite{rectifiy_flow}, treating the latent depth as the temporal dimension.
Let $\mat{Z}_{\epsilon} \sim \mathcal{N}(0,\mat{I})$ be a standard Gaussian noise sample. For a timestep $t\in[0,1]$, the forward process is defined as:
\begin{align} 
    \label{eq:interpolation}
    \mat{Z}_t = t\mat{Z}_{\text{CE}} + (1-t)\mat{Z}_{\epsilon}.
\end{align}
The ground-truth velocity field driving the transport from noise to data is $\mat{V}_t = \mat{Z}_{\text{CE}} - \mat{Z}_{\epsilon}$.
The DiT model $v_\theta$ learns to predict this velocity conditioned on $\mat{Z}_{\text{NC}}$ and $t$, minimizing the flow matching objective:
\begin{align}
    \label{eq:fm_loss}
    \mathcal{L}_{\text{FM}} = \mathbb{E}_{\mat{Z}_{\epsilon}, \mat{Z}_{\text{CE}}, \mat{Z}_{\text{NC}}, t}
    \left[ \| v_\theta(\mat{Z}_t, \mat{Z}_{\text{NC}}, t) - \mat{V}_t \|_2^2 \right].
\end{align}
By formulating CT volumes as latent token sequences, \methodname{} synthesizes multiple slices jointly and captures long-range anatomical dependencies.
To ensure anatomy-phase consistent synthesis, we further introduce \modelname{} and \alignname{}, detailed below.

\subsection{Anatomy-Routed Mixture-of-Experts (\modelname{})}

Different anatomical structures exhibit distinct enhancement dynamics---for instance, the liver and kidney follow different contrast uptake patterns.
Modeling such anatomical heterogeneity is challenging for latent diffusion models, as their representations primarily capture low-level textures rather than high-level semantics~\cite{rae,svg}.
While auxiliary segmentation objectives can improve anatomical awareness~\cite{zongwei_ce2ct, aldm}, they add substantial computational overhead due to the extra VAE decoding step and multi-task optimization.
To address this, we propose \modelname{}, which leverages readily available organ segmentation as explicit routing guidance to decouple heterogeneous enhancement patterns in a divide-and-conquer manner, anchoring each pattern to its corresponding anatomical semantics.

\paragraph{Anatomy router.}
Given an input NCCT volume $\mat{X}_{\text{NC}}$, we first obtain a dense segmentation map using TotalSegmentor~\cite{totalsegmentator}, and then select $K$ organs with distinct enhancement dynamics for routing.
The segmentation map is downsampled spatially and temporally to align with the latent token sequence, yielding token-wise organ labels $s_n \in\{0, 1,\dots,K\}$ for each token $n \in\{1,\dots,N\}$, where $s_n = 0$ indicates non-selected organs. 
We define the binary anatomy mask $\mat{S}_k \in \{0,1\}^{N}$ for the $k$-th organ, where $[\mat{S}_k]_n = \mathbf{1}[s_n = k]$.
The routing function aggregates outputs from organ-specific experts, while tokens with $s_n = 0$ remain unchanged:
\begin{align}
    \label{eq:moe}
    \mat{F}_{\text{out}} = \mat{F}_{\text{in}} + \sum_{k=1}^{K} E_{k}\!\left(\mat{F}_{\text{in}}, t, \mat{S}_k\right),
\end{align}
where $\mat{F}_{\text{in}}, \mat{F}_{\text{out}} \in \mathbb{R}^{N \times C}$ denote the input and output features, and $E_k(\cdot)$ is the expert specialized for the $k$-th organ.

\paragraph{Organ-specific expert.}
Each expert $E_k$ consists of two branches, both gated by the anatomy mask $\mat{S}_k$ to ensure anatomy-specific processing.
The first branch applies a standard FFN for token-wise feature refinement.
The second branch leverages a learnable organ-specific memory $\mat{M}_k \in \mathbb{R}^{L \times C}$ to capture discriminative enhancement patterns. 
Similar to Flamingo~\cite{flamingo}, the memory entries are learnable parameters that serve as queries to aggregate salient features from input tokens.
Let $\mat{Q} = \mat{M}_k \mat{W}_q \in \mathbb{R}^{L \times C}$, $\mat{K} = \mat{F}_{\text{in}} \mat{W}_k \in \mathbb{R}^{N \times C}$, and $\mat{V} = \mat{F}_{\text{in}} \mat{W}_v \in \mathbb{R}^{N \times C}$. 
The attention weights $\mat{A} = \text{softmax}(\mat{Q}\mat{K}^\top / \sqrt{C}) \in \mathbb{R}^{L \times N}$ first aggregate input tokens into $L$ memory slots, then redistribute back to each token position as $\mat{H} = \mat{A}^\top (\mat{A} \mat{V}) \in \mathbb{R}^{N \times C}$.
Both branches are modulated by time-dependent gates $\mathcal{G}_k^{\text{ffn}}(t), \mathcal{G}_k^{\text{mem}}(t) \in \mathbb{R}^{C}$. The output of expert $E_k(\mat{F}_{\text{in}}, t, \mat{S}_k)$ is:
\begin{align}
    \label{eq:expert_internal}
    \mat{S}_k \odot \Big( 
        \mathcal{G}_k^{\text{ffn}}(t) \odot \text{FFN}_k(\mat{F}_{\text{in}}) 
        + \mathcal{G}_k^{\text{mem}}(t) \odot \mat{H} \Big).
\end{align}

%  ========== Dataset Statistics =======
%  ========== Dataset Statistics =======
\begin{table*}[t]
\caption{Statistics of the datasets used in this study. The Abdomen denotes an in-house dataset collected from the First Affiliated Hospital, Zhejiang University School of Medicine. ``Total (\#)'' refers to the number of raw cases, while ``Total'' indicates the number of valid cases after data cleaning and registration.}
\label{tab:datasets}
\centering
\small
\renewcommand{\arraystretch}{1.1}
\setlength{\tabcolsep}{4pt}
\begin{tabular}{l c c c c c c c} 
\shline
Dataset 
% & Region 
& Total (\#) 
& Total 
& Thickness (mm) 
& Slice Range 
& Train (vol./slices) 
& Test (vol./slices) \\
\hline

% 【】
VinDr-Multiphase~\cite{dataset_vindr} 
% & Abdomen 
& 183 
& 168 
& [0.3, 5.0] 
& [38, 1745] 
& 130 / 320,140 
& 38 / 20,905 \\

% 【】
WAW-TACE~\cite{wawtrace}          
% & Abdomen 
& 233 
& 159 
& [0.8, 7.5] 
& [36, 526] 
& 138 / 91,225 
& 21 / 16,890 \\

% 【】
Abdomen
% & Abdomen 
& 900 
& 900 
& [0.45, 5.0]
& [42, 540] 
& 800 / 373,647 
& 100 / 46,519 \\
\shline
\end{tabular}
\vspace{-6pt}
\end{table*}

\begingroup
\color{blue}
\begin{table}[t]
    \caption{Organ groupings for routing.}
    \label{tab:organ_groups}
    \centering
    \small
    \begin{tabular}{p{2.5cm}p{5cm}}
    \shline
    Routing Group & Organs \\
    \hline
    Enhancing & liver, spleen, pancreas, kidney, adrenal gland, gallbladder, stomach, esophagus, small bowel, duodenum, colon, urinary bladder, prostate \\
    Parenchymal & liver, spleen, pancreas, kidney, prostate \\
    Motion-stable & liver, spleen, gallbladder, pancreas, kidney, adrenal gland, esophagus, prostate, urinary bladder \\
    \shline
    \end{tabular}
    \vspace{-4pt}
\end{table}
\endgroup

\subsection{Intensity-Phase Aware Representation Alignment (\alignname)}

High-fidelity VCE requires accurately synthesizing subtle contrast-induced signals, yet this is hindered by two factors.
First, CT spans a dynamic range exceeding 2000 HU, while contrast enhancement is typically limited to 20-50 HU, making these signals easily overshadowed.
Second, spatial misalignment inevitably persists even after deformable registration due to respiration and organ motion, introducing intensity discrepancies that corrupt phase-specific contrast details.
As a result, learning a model with conventional objectives may entangle true enhancement signals with registration noise.

To ensure phase-consistent enhancement, we propose intensity-phase aware representation alignment (\alignname), which steers the model toward fine-grained contrast cues while mitigating the impact of spatial misalignment.
Given paired NCCT and CECT volumes $\mat{X}_{\text{NC}}$ and $\mat{X}_{\text{CE}}$, we first apply a soft-tissue HU window of $[-175, 275]$ to emphasize enhancement-relevant intensity ranges.
A pre-trained teacher model then extracts slice-wise features, followed by temporal and spatial downsampling consistent with the DiT token sequence, producing $\mathbf{T}_{\text{NC}}, \mathbf{T}_{\text{CE}} \in \mathbb{R}^{N \times C_t}$.
For alignment, we project the DiT latent features into the teacher space via an MLP layer, yielding $\mathbf{H} \in \mathbb{R}^{N \times C_t}$.

\paragraph{Intensity-based weighting.} 
To emphasize contrast-enhanced regions while down-weighting misaligned areas, we introduce a token-wise weighting scheme based on intensity differences.
Specifically, we compute the voxel-wise difference map $\Delta \mat{X} = \mat{X}_{\text{CE}} - \mat{X}_{\text{NC}}$, and assign importance based on the physical prior that contrast agents increase tissue attenuation:
\begin{equation}
    \mat{W}_{p} =
    \begin{cases}
        \alpha_{\text{enh}}, & \text{if } \Delta \mat{X}_{p} > \tau_{\text{pos}}, \\
        \alpha_{\text{mis}}, & \text{if } \Delta \mat{X}_{p} < 0, \\
        \alpha_{\text{bg}},  & \text{otherwise},
    \end{cases}
    \label{eq:token_weight}
\end{equation}
where $p$ indexes spatial locations. Positive differences indicate true enhancement, while negative differences typically arise from misalignment.
We downsample $\mat{W}$ to align with the latent token sequence, yielding token-level weights $\mathbf{w} \in \mathbb{R}^{N}$. 
% \longzili{add tau pos setting}
By default, we set $\tau_{\text{pos}}=20$ HU, following the common CT criterion for definite enhancement.

\paragraph{Intensity-aware alignment.}
We model contrast enhancement as a residual signal in the teacher feature space, and align the student representation to this residual. 
Let $\mathrm{N}(\cdot)$ denote channel-wise $L_2$ normalization:
\begin{equation}
    \mathcal{L}_{\text{int}}
    =
    \left\|
    \mathbf{w}\odot
    \left(
        \mathrm{N}(\mathbf{H})
        -
        \mathrm{N}(\mathbf{T}_{\text{CE}}-\mathbf{T}_{\text{NC}})
    \right)
    \right\|_1.
\end{equation}
This objective guides the model to focus on phase-specific enhancement signals rather than registration noise or anatomical structure.

\paragraph{Phase-aware alignment.}
Token-wise alignment alone cannot guarantee global structural consistency. Since contrast enhancement is physically anchored to the underlying anatomy, the synthesized features should respect this structural dependency.
To enforce this, we align the student's internal token relationships with the teacher's cross-phase correspondence, which can be written as:
\begin{equation}
    \mathcal{L}_{\text{phase}}
    \!=\!
    \left\|
     \mathbf{w}\mathbf{w}\!^\top \!\odot\! \left( 
        \mathrm{N}(\mathbf{H})\mathrm{N}(\mathbf{H})\!^\top 
        \!-\! 
        \mathrm{N}(\mathbf{T}_{\text{NC}})\mathrm{N}(\mathbf{T}_{\text{CE}})\!^\top\!
     \right) 
    \right\|_1. 
\end{equation}
The term $\mathrm{N}(\mathbf{T}_{\text{NC}})\mathrm{N}(\mathbf{T}_{\text{CE}})^\top$ captures the cross-phase correspondence from the teacher, 
encoding how anatomical structures in NCCT relate to enhancement patterns in CECT. By aligning the student's self-similarity to this map, we anchor the synthesis to the physical causality of contrast enhancement.

\subsection{Objective Function}
The final objective combines the flow matching loss with the proposed alignment terms:
\begin{align}
    \label{eq:total_loss}
    \mathcal{L}_{\text{total}}
    =
    \mathcal{L}_{\text{FM}}
    + \lambda_{\text{int}}\mathcal{L}_{\text{int}}
    + \lambda_{\text{phase}}\mathcal{L}_{\text{phase}},
    % + \lambda_{\text{struct}}\mathcal{L}_{\text{struct}},
\end{align}
where $\lambda_{\text{int}}$ and $\lambda_{\text{phase}}$ are hyperparameters to balance intensity- and phase-aware alignment, respectively.

 % radar
\begin{figure*}[t]
    \centering    
    \includegraphics[width=0.85\linewidth]{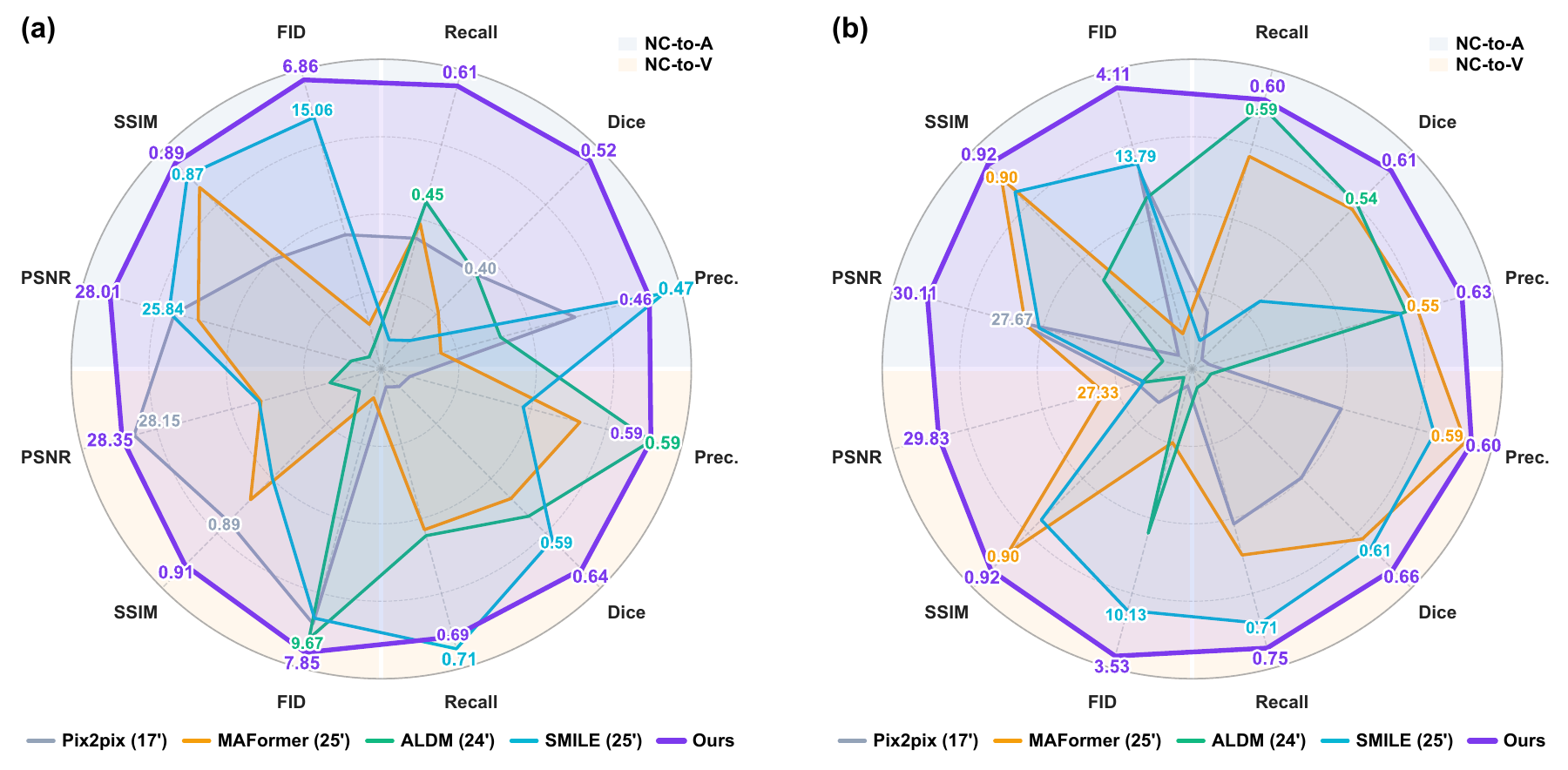}
    \caption{Quantitative comparison of NCCT-to-CECT synthesis on the (a) VinDr-Multiphase and (b) WAW-TRACE datasets. ``Prec.'' denotes the precision score. }
    % For visualization clarity, only a subset of comparative methods is plotted.}
    % \vspace{-8pt}
    % \vspace{-6pt}
    \label{fig:phaser_main_result} 
\end{figure*}
%  radar

\begin{table*}[t]
    \caption{Quantitative evaluation for state-of-the-art methods on Abdomen dataset. The best results are marked in \textbf{bold}.}
    \setlength{\tabcolsep}{5pt}
    \small
    \label{tab:results_inhouse}
    \renewcommand{\arraystretch}{1.0}
    \centering
{
\begin{tabular}{lcccccc|cccccc}
\shline
\multirow{2}{*}{Methods} & \multicolumn{6}{c|}{NC$\rightarrow$A}
& \multicolumn{6}{c}{NC$\rightarrow$V} \\

& PSNR & SSIM & Dice & Recall & Prec. & FID
& PSNR & SSIM & Dice & Recall & Prec. & FID \\
\hline

CycleGAN~\cite{cyclegan}
& 29.29 & 0.78 & 0.20 & 0.38 & 0.14 & 32.17
& 29.30 & 0.83 & 0.42 & 0.47 & 0.39 & 27.22 \\

Pix2pix~\cite{pix2pix}
& 29.54 & 0.84 & 0.16 & 0.17 & 0.16 & 20.79
& 30.65 & 0.93 & 0.57 & 0.55 & 0.59 & 10.76 \\

EaGAN~\cite{eagan}
& 23.47 & 0.59 & 0.29 & 0.35 & 0.26 & 25.97
& 28.89 & 0.82 & 0.38 & 0.44 & 0.34 & 22.88 \\

MAFormer~\cite{maformer}
& 29.48 & 0.93 & 0.35 & 0.37 & 0.33 & 15.49
& 28.76 & 0.93 & 0.60 & 0.68 & 0.54 & 16.57 \\

ALDM~\cite{aldm}
& 30.72 & 0.92 & 0.40 & 0.51 & 0.33 & 10.66
& 27.26 & 0.89 & 0.41 & 0.33 & 0.54 & 3.25 \\

SMILE~\cite{zongwei_ce2ct}
& 26.90 & 0.87 & 0.25 & 0.22 & 0.31 & 9.12
& 26.61 & 0.86 & 0.54 & 0.69 & 0.44 & 9.21 \\

WAN 2.1~\cite{wan21}
& 32.43 & 0.95 & 0.50 & 0.47 & 0.53 & 1.68
& 31.72 & \textbf{0.95} & 0.66 & 0.63 & \textbf{0.71} & 1.35 \\
\rowcolor{cyan!10}
\methodname
& \textbf{33.19} & \textbf{0.96} & \textbf{0.55} & \textbf{0.57} & \textbf{0.54} & \textbf{0.99}
& \textbf{32.71} & \textbf{0.95} & \textbf{0.71} & \textbf{0.70} & \textbf{0.71} & \textbf{0.94} \\

\shline
\end{tabular}
}
\vspace{-6pt}
\end{table*}
%  ========== Dataset Statistics =======
%  ========== Dataset Statistics =======
%  ========== Dataset Statistics =======
%  ========== Dataset Statistics =======

%% file: secs/result.tex
% ===========

\section{Result}

\subsection{Experimental Setup}

% ======================
% ======================

\paragraph{Datasets.}
As summarized in Table~\ref{tab:datasets}, we evaluate \methodname{} on VinDr-Multiphase~\cite{dataset_vindr} and WAW-TACE~\cite{wawtrace} dataset, alongside a large-scale abdomen dataset collated from the First Affiliated Hospital, Zhejiang University School of Medicine. 
Given the frequent absence of the Delayed phase, we restrict our study to the Non-contrast (NC), Arterial (A), and Portal Venous (V) phases, excluding subjects with incomplete data. 
For preprocessing, we resample the A and V volumes to the NC geometry using linear interpolation via SimpleITK~\cite{simpleitk}, padding out-of-field voxels to -1024 HU.  
Unless noted otherwise, we employ DEEDS~\cite{deeds1, deeds2} to spatially register the contrast phases to the NC reference. 
Finally, the data is split into training and validation sets at an approximate 8:1 ratio without subject overlap.

\paragraph{Implementation details.}
Our framework is built upon the WAN 2.1~\cite{wan21}, utilizing a 1.3B-parameter DiT backbone, and a VAE fine-tuned on CT data. 
First, to bridge the domain gap between video and CT images, we initialize the DiT from WAN 2.1 and perform warm-up pretraining on the AbdomenAtlas 3.0~\cite{abdomen_atlas3} dataset for 50 epochs, with the VAE encoder and decoder, as well as the T5 text encoder frozen. We set the model in a text-conditioned generation mode with the prompt ``\texttt{generate CT volume in phase \{\}}''. This step is necessary for stable training due to the large domain gap between natural videos and CT volumes.
Second, we initialize the DiT using checkpoint from the first stage, integrate \modelname{} into the DiT, and continue fine-tuning for conditional synthesis with NCCT as input for 24k iterations. Following WAN 2.1, we simply concatenate the NCCT volume to the model input. The prompt is set to ``\texttt{convert CT volume from phase \{\} to phase \{\}}''. We empirically set $\lambda_{\text{int}}=0.1$ and $\lambda_{\text{phase}}=1$, making them approximately one-tenth of the diffusion loss. For \alignname{}, we apply it to the 8-th block of the DiT, as this choice is well-supported by prior work~\cite{repa, videorepa}.
For both stages, each sample is augmented via random spatial cropping to $448 \times 448$ followed by resizing to $512 \times 512$, along with random flipping. This augmentation strategy is essential because multi-phase paired data is relatively scarce, and increasing data diversity significantly improves training stability. 
The organ groups for the anatomy router in our experiments are shown in Table~\ref{tab:organ_groups}, and we use the parenchymal group by default. 
We optimize \methodname{} using Adam with $\beta_1=0.9$, $\beta_2=0.99$, and a learning rate of $1\times 10^{-4}$, with a global batch size of 16 on 8 NVIDIA H20 GPUs.

% [使用一个混合的图]
\begin{figure*}[t]
    \centering
    \includegraphics[width=0.95\linewidth]{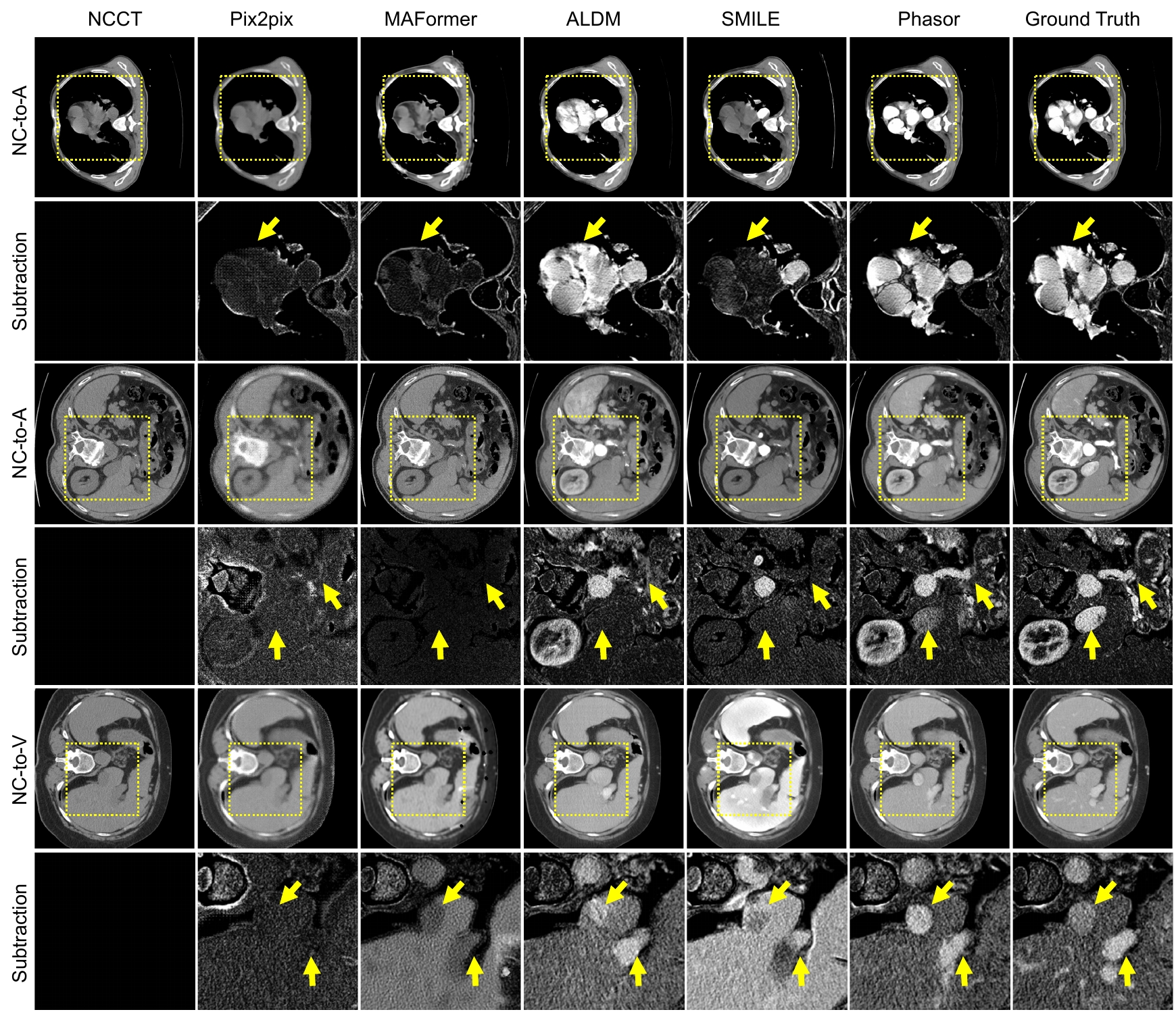}
    \caption{Qualitative comparison of different methods on the VinDr-Multiphase and WAW-TRACE dataset. The display window is [-160, 240] HU.}
    \vspace{-8pt}
    \label{fig:visual_compare_main} 
\end{figure*}

\begin{figure*}[t]
    \centering
    \includegraphics[width=0.95\linewidth]{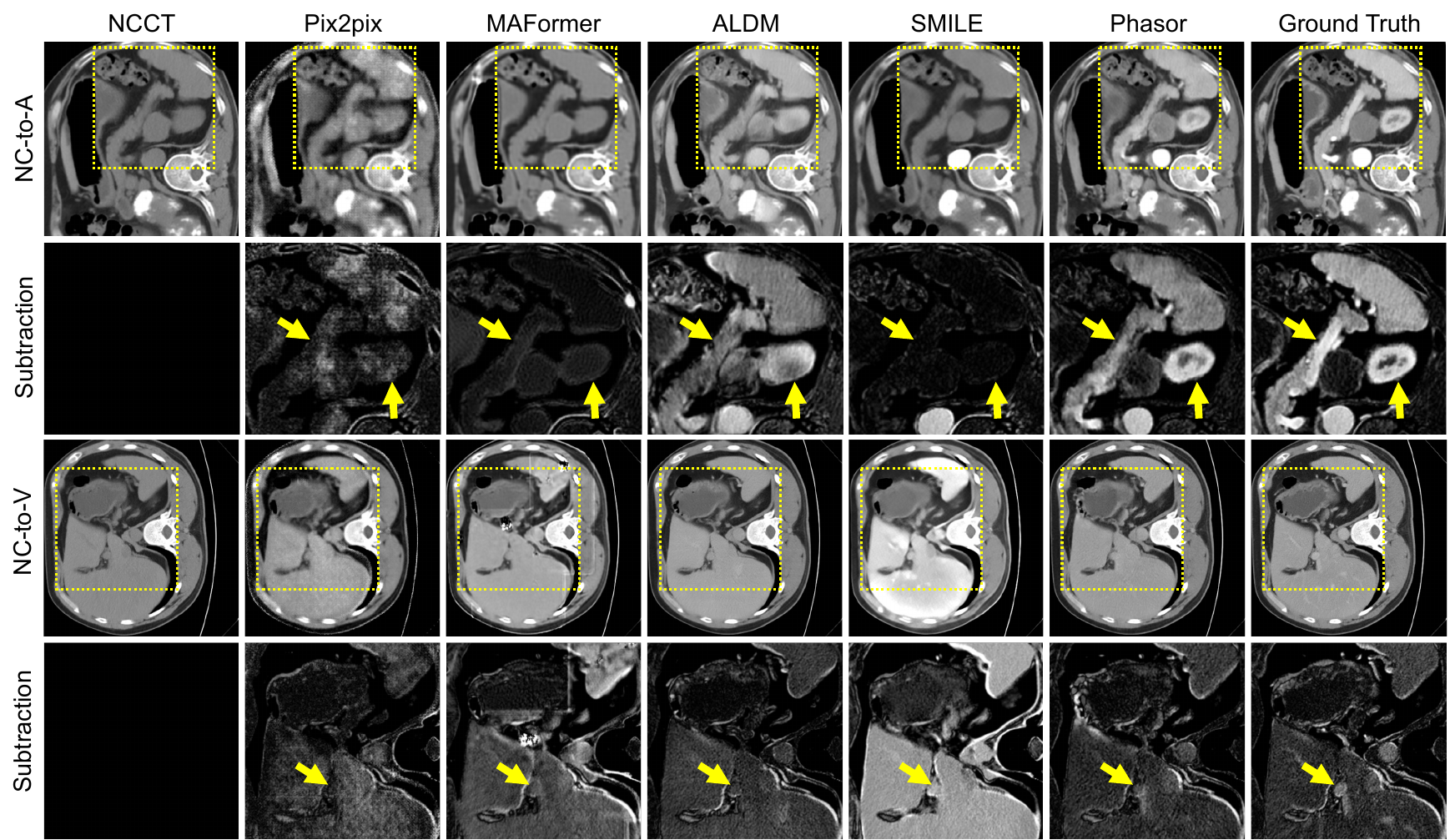}
    \caption{Qualitative comparison of different methods on the Abdomen dataset. The display window is [-160, 240] HU.}
    \vspace{-8pt}
    \label{fig:visual_compare_inhouse}
\end{figure*}

\begin{figure}[t]
    \centering    
    \includegraphics[width=1.0\linewidth]{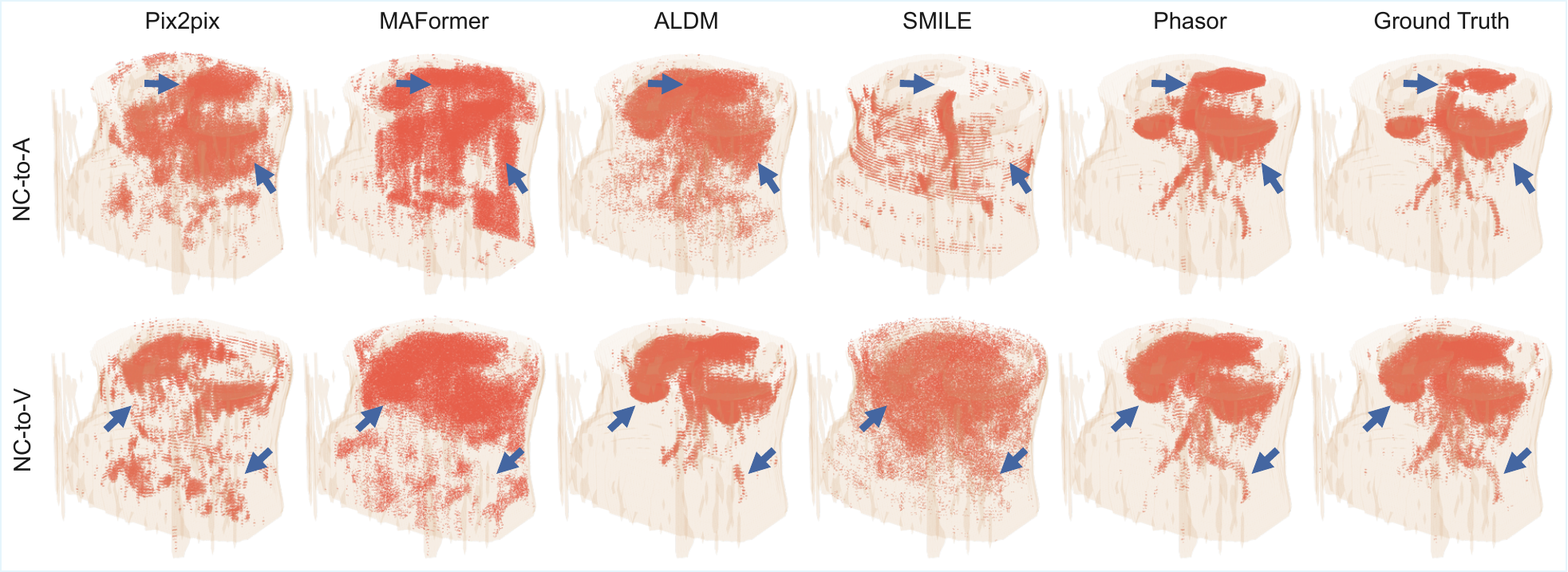}
    \caption{3D visualization of the subtraction volume between the synthesized CECT and the NCCT. Zoom in for better view.}
    \vspace{-8pt}
    \label{fig:visual_compare_3d} 
\end{figure}

\paragraph{Comparison methods.}
We compare \methodname{} with a comprehensive set of baselines: The GAN-based 3D-CycleGAN~\cite{cyclegan} and 3D-Pix2Pix~\cite{pix2pix}.
MAFormer~\cite{maformer}, a state-of-the-art 2D model utilizing structure-invariant losses for fine-grained details; and volumetric diffusion models including ALDM~\cite{aldm} and WAN~\cite{wan21}.  
We also include SMILE~\cite{zongwei_ce2ct}, a recent diffusion model tailored for precise contrast enhancement. 
For SMILE, we employ the official checkpoint; for all other methods, given the absence of official weights for CT VCE, we retrain them on our dataset to ensure a fair comparison.

\paragraph{Metrics.}
We evaluate synthesis quality from two perspectives: \emph{reconstruction fidelity} and \emph{enhancement accuracy}. For general image quality, we report Peak Signal-to-Noise Ratio (PSNR) and 3D Structural Similarity Index Measure (3D-SSIM)~\cite{ssim} to evaluate pixel-level reconstruction accuracy and perceptual structural similarity, respectively. Additionally, we utilize Fréchet Inception Distance (FID)~\cite{fid} to measure the distributional distance between the synthesized and ground-truth volumes. 
To quantify enhancement accuracy, we compute residual maps $\Delta \mat{X} = \mat{X}_{\text{CE}} - \mat{X}_{\text{NC}}$ and $\Delta \widehat{\mat{X}} = \widehat{\mat{X}}_{\text{CE}} - \mat{X}_{\text{NC}}$, and threshold at 20 HU to obtain binary enhancement masks $\mathbf{E}$ and $\widehat{\mathbf{E}}$ for the ground truth and the synthesized ones, respectively.
We then report Dice~\cite{dice}, Recall, and Precision between $\widehat{\mathbf{E}}$ and the ground truth $\mathbf{E}$ to evaluate spatial overlap and sensitivity.

\subsection{Comparison with State-of-the-Art Methods}
%  【1】
\paragraph{Quantitative results on Vindr-Multiphase and WAW-Trace dataset.}
Fig.~\ref{fig:phaser_main_result} presents the quantitative results of each method. 
On the left side of the radar chart, the image fidelity metrics indicate that GAN-based methods remain competitive with diffusion models. Notably, MAFormer (orange curve), despite its 2D synthesis framework, achieves strong SSIM scores, highlighting the efficacy of structure-aware modeling in preserving anatomical details. In contrast, Pix2pix exhibits instability and underperforms across most metrics.

On the right side of the radar chart, the enhancement accuracy metrics indicate a clear advantage of diffusion-based methods for enhancement accuracy. Specifically, SMILE performs particularly well on the NC-to-V synthesis but shows reduced accuracy in NC-to-A synthesis. This limitation may stem from its 2D formulation, which struggles to model the longitudinal continuity of vascular structures critical for arterial enhancement.
Overall, the proposed \methodname{} consistently outperforms all baselines across datasets, achieving especially pronounced gains in enhancement accuracy while maintaining competitive synthesis fidelity.

\paragraph{Quantitative results on abdomen dataset.}
Table~\ref{tab:results_inhouse} summarizes the results on abdomen dataset. Given the large scale of this training set, these results better reflect the upper-bound performance capabilities of each method.

First, we observe that with well-registered data and sufficient training samples, GAN-based methods can achieve good results on low-level metrics such as PSNR. Specifically, Pix2pix attains PSNR scores comparable to diffusion-based models. While MAFormer is less competitive on low-level metrics due to its slice-to-slice formulation, we find that 3D GANs perform poorly on enhancement accuracy. In contrast, MAFormer achieves substantially higher enhancement accuracy than 3D GANs, suggesting that conventional low-level metrics may not faithfully reflect the correctness of contrast enhancement.

Second, comparing diffusion-based models reveals that volumetric approaches have clear advantages over 2D methods, particularly for the arterial phase, where the contrast agent propagates along arterial structures. This indicates that jointly modeling multiple slices is beneficial. While the baseline WAN achieves a relatively balanced trade-off between image quality and enhancement accuracy, \methodname{} outperforms it across all metrics by significant margins. 

Third, consistent trends are observed when comparing results from the non-contrast phase to the arterial and venous phases. Since we directly utilize the official checkpoint of SMILE, it underperforms on the arterial phase while remaining good performance on the venous phase. Overall, \methodname{} is robust across phases, achieving leading enhancement accuracy with high synthesis fidelity.

\subsection{Visual Comparsion}
Fig.~\ref{fig:visual_compare_main} and Fig.~\ref{fig:visual_compare_inhouse} present the visual results alongside subtraction maps, which are commonly used in clinical practice to visualize contrast enhancement patterns.
First, while GAN-based models enhance the images to some extent with good reconstruction fidelity, they exhibit considerable errors in the subtraction maps. 
Diffusion model ALDM and WAN achieve improved image quality, but also suffer from high error rates in their subtraction maps; for instance, ALDM generates excessive contrast agent, whereas WAN generates too little. This suggests that CT VCE demands substantially higher fidelity than simple style transfer. 
We also find that SMILE achieves relatively high image quality but tends to be conservative in contrast-enhancing regions.
In contrast, \methodname achieves high image quality and produces subtraction maps that are highly similar to the ground truth. 

\begin{table}[t]
    \caption{Cross-dataset generalization from the Abdomen dataset to the WAW-TRACE dataset without any fine-tuning.}
    \setlength{\tabcolsep}{5pt}
    \small
    \label{tab:results_wawtace_generalization}
    \renewcommand{\arraystretch}{1.2}
    \centering
{
\begin{tabular}{lccc|ccc}
\shline
\multirow{2}{*}{Methods} 
& \multicolumn{3}{c|}{NC$\rightarrow$A}
& \multicolumn{3}{c}{NC$\rightarrow$V} \\

& PSNR & Dice & FID
& PSNR & Dice & FID \\
\hline

MAFormer~\cite{maformer} & 25.22 & 0.33 & 44.30 & 24.33 & 0.49 & 45.24 \\
ALDM~\cite{aldm}        & 22.49 & 0.43 & 27.25 & 24.35 & 0.19 & 26.47 \\
SMILE~\cite{zongwei_ce2ct} & 27.11 & 0.34 & 13.79 & 26.73 & \textbf{0.61} & 10.13 \\
\rowcolor{cyan!4}
\methodname{}             
& \textbf{29.04} & \textbf{0.53} & \textbf{6.27} 
& \textbf{29.48} & 0.58 & \textbf{4.85} \\

\shline
\end{tabular}
}
\vspace{-8pt}
\end{table}

Fig.~\ref{fig:visual_compare_3d} presents 3D visualizations of the enhanced regions for each method. \methodname{} achieves a distribution similar to the ground truth, particularly for vertical structures such as blood vessels. Compared to ALDM and SMILE, \methodname{} generates results that are more realistic in terms of both overall structure and details.

\subsection{Cross-dataset Generalization}
Table~\ref{tab:results_wawtace_generalization} presents the results of models trained on the Abdomen dataset and directly evaluated on the WAW-TACE dataset without any fine-tuning. 
We observe that \methodname{} maintains strong cross-dataset generalization ability.  In addition, \methodname{} consistently outperforms the competing methods by 2-7 dB in synthesis quality.

\begin{table*}[t]
\caption{
Ablation of components in \methodname{} on the VinDr-Multiphase dataset.
We report results with and without deformable registration.
All models are trained for 3.2k iterations.
The best and second-best results are highlighted in \textbf{bold} and \underline{underline}, respectively.
}
\scriptsize
\label{tab:ablation}
\renewcommand{\arraystretch}{1.0}
\centering
\resizebox{\textwidth}{!}{
\begin{tabular}{ccc | cccccc | cccccc}
% \begin{tabular}{ccc | rrrrrr | rrrrrr}
\shline
\multicolumn{3}{c|}{} & \multicolumn{6}{c|}{Unregistered training data} & \multicolumn{6}{c}{Registered training data} \\
\cline{4-15}
% \cline{4-9} 
% \cline{10-15}

3D & \modelname{} & \alignname &
PSNR & SSIM & Dice & Recall & Prec. & FID &
PSNR & SSIM & Dice & Recall & Prec. & FID \\
\hline

 &  &
& 16.447 & 0.181 & 0.296 & \textbf{0.700} & 0.193 & 62.537
& 19.484 & 0.540 & 0.390 & 0.407 & 0.383 & 32.537 \\

$\checkmark$ &  &
& 23.989 & 0.803 & 0.483 & \underline{0.640} & 0.397 & 10.656
& 27.160 & 0.882 & 0.482 & 0.422 & \textbf{0.573} & \underline{7.694} \\

$\checkmark$ & $\checkmark$ &
& \textbf{25.203} & \textbf{0.826} & 0.492 & 0.520 & \underline{0.483} & \underline{8.364}
& \underline{27.665} & \textbf{0.895} & 0.539 & \underline{0.591} & 0.501 & 7.735 \\

$\checkmark$ &  & $\checkmark$
& 24.926 & \underline{0.824} & \underline{0.507} & 0.548 & \textbf{0.487} & 8.662
& 27.237 & 0.888 & \underline{0.540} & \textbf{0.595} & 0.500 & 7.923 \\

$\checkmark$ & $\checkmark$ & $\checkmark$
& \underline{25.034} & \underline{0.824} & \textbf{0.508} & 0.578 & 0.464 & \textbf{8.269}
& \textbf{27.734} & \underline{0.894} & \textbf{0.541} & 0.553 & \underline{0.536} & \textbf{7.653} \\

\shline
\end{tabular}
}
\vspace{-6pt}
\end{table*}

\begin{figure*}[t]
    \centering    
    \includegraphics[width=1.0\linewidth]{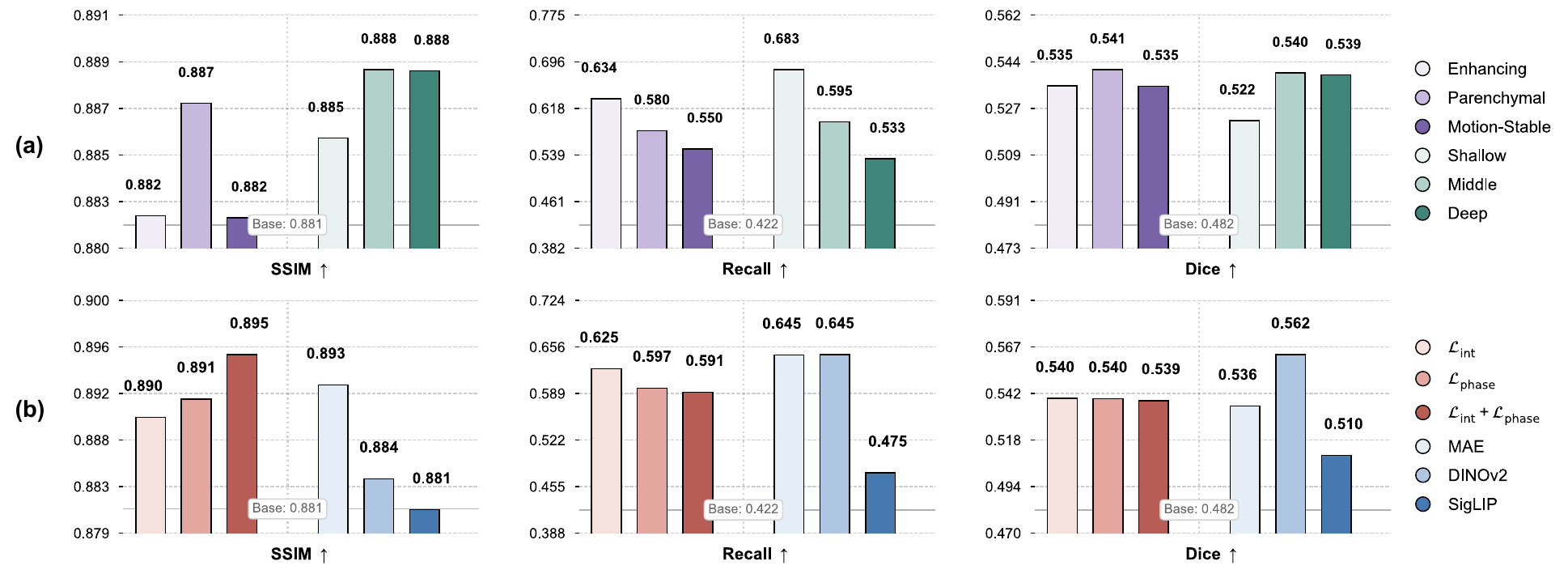}
    \caption{
    Detailed ablation studies of the design of \methodname{}. (a) Ablation study on the design of \modelname{}, including the impact of selecting different anatomical structures for the routing mechanism, and MoE placement by inserting \modelname{} layers at different depths---shallow (layers [2,4,8]), middle (layers [14,16,18]), and deep (layers [24,26,28])---within the 30-layer DiT model. 
    (b) Ablation on \alignname, using different losses and teacher models.
    } 
    \label{fig:ablation_detail} 
    % \vspace{-4pt}
\end{figure*}

\subsection{Ablation Study}

We aim to answer three questions in the ablation study:
(1) the impact of each design of \methodname{},
(2) how to configure an effective MoE, and
(3) which alignment strategies are most helpful for VCE.
All experiments are conducted on the VinDr-Multiphase dataset for the NC-to-A synthesis. We train each model for 3.2k iterations using the same settings.

% =========== 

\paragraph{Ablation on components in \methodname.}
Table~\ref{tab:ablation} summarizes the ablation results of \methodname{}. To study the effect of volumetric modeling, we compare a 2D variant that processes a single slice at each step, while the 3D variant takes 9 slices. For the dataset without deformable registration, we directly utilize the data after resampling to the NC geometry. This is a commonly used preprocessing strategy that aligns volumes at the spatial level but retains noticeable pixel-level misalignment. We use registered data for evaluation.

%  ====== 
% 
By analyzing the results, we have 4 findings. 
\underline{\textbf{(i)}} Registration is critical.
While deformable registration is time-consuming and requires extra preprocessing, some prior work trains on unregistered data~\cite{zongwei_ce2ct, ozbey2023unsupervised}. However, our results show that registration is essential: training with registered data yields substantial PSNR gains and benefits both 2D and 3D models. Otherwise, the results show inflated recall but lower Dice, suggesting that models tend to hallucinate enhancement regions to reduce the loss. This supports the use of high-quality alignment as a key step in building reliable VCE pipelines.
\underline{\textbf{(ii)}} 3D modeling improves VCE.
We observe that the 3D variant consistently outperforms the 2D counterpart, indicating that multi-slice context substantially improves enhancement accuracy, regardless of whether the training data are registered. 
\underline{\textbf{(iii)}} \alignname{} boosts detail fidelity, consistently improving low-level metrics across both unregistered and registered datasets. This confirms that mitigating spatial-intensity conflicts enables the model to focus on contrast details. Notably, the gains are more pronounced on unregistered data, likely due to the re-weighting mechanism in Eq.~\ref{eq:token_weight} that effectively down-weights misaligned tokens while highlighting contrast regions. \underline{\textbf{(iv)}} \modelname{} better benefit the enhancement accuracy. It effectively decouples the learning of heterogeneous tissue patterns, ensuring that distinct anatomical structures are enhanced with high specificity. 
Combining both strategies, \methodname achieves marked improvements on both synthesis fidelity and accuracy. 
In summary, each component plays an important role in \methodname{}, and our findings provide practical guidance for future works.

\paragraph{Ablation on \modelname{}.}
Fig.~\ref{fig:ablation_detail}(a) present detailed ablations of \modelname{}.
First, we compare routing strategies based on three organ groupings: enhancing tissues, parenchymal organs (\eg, liver and pancreas), and motion-stable organs, as listed in Table~\ref{tab:organ_groups}. 
Routing via parenchymal groups achieves the best SSIM and Dice. By contrast, groups containing hollow or highly mobile organs (\eg, colon) show limited improvement.
Second, we analyze the insertion depth of \modelname{} within the DiT backbone. Placing \modelname{} in shallow layers yields the highest recall, indicating that enhancement regions are strongly organ-dependent. However, insertion in the middle layers provides the optimal balance with the highest Dice, while insertion in deep layers achieves worse results. This suggests that VCE relies on both the high-level semantics of where to enhance and the low-level details of how to synthesize textures, requiring a design that balances both.

\paragraph{Ablation on \alignname.}
We analyze the detailed design choices of \alignname in Fig.~\ref{fig:ablation_detail}(b).
First, we evaluate the contributions of losses $\mathcal{L}_{\text{int}}$ and $\mathcal{L}_{\text{phase}}$. Both terms individually improve synthesis quality over the baseline. Combining them yields the best performance, achieving a notable boost in both PSNR and Dice metrics. This suggests that feature distillation from a teacher model provides effective guidance on fine-grained textures, which is crucial for synthesis fidelity. 
Second, we compare different pretrained encoders as the teacher model. MAE~\cite{MAE} outperforms other candidates, likely because its reconstruction-based self-supervised objective captures subtle intensity variations inherent to CT images, consistent with prior findings~\cite{repa, rae, dinov3bench}. In contrast, DINOv2~\cite{dinov2} and SigLIP~\cite{siglip} are pretrained on natural images and are less sensitive to these intensity nuances. Given the limited availability of CT-specific pretrained encoders, we adopt MAE as the default teacher to enhance details.

% \paragraph{More Ablation.}
\paragraph{Ablation on the number of organ-specific memories.}
\begin{figure}[t]
    \centering
    \includegraphics[width=0.9\linewidth]{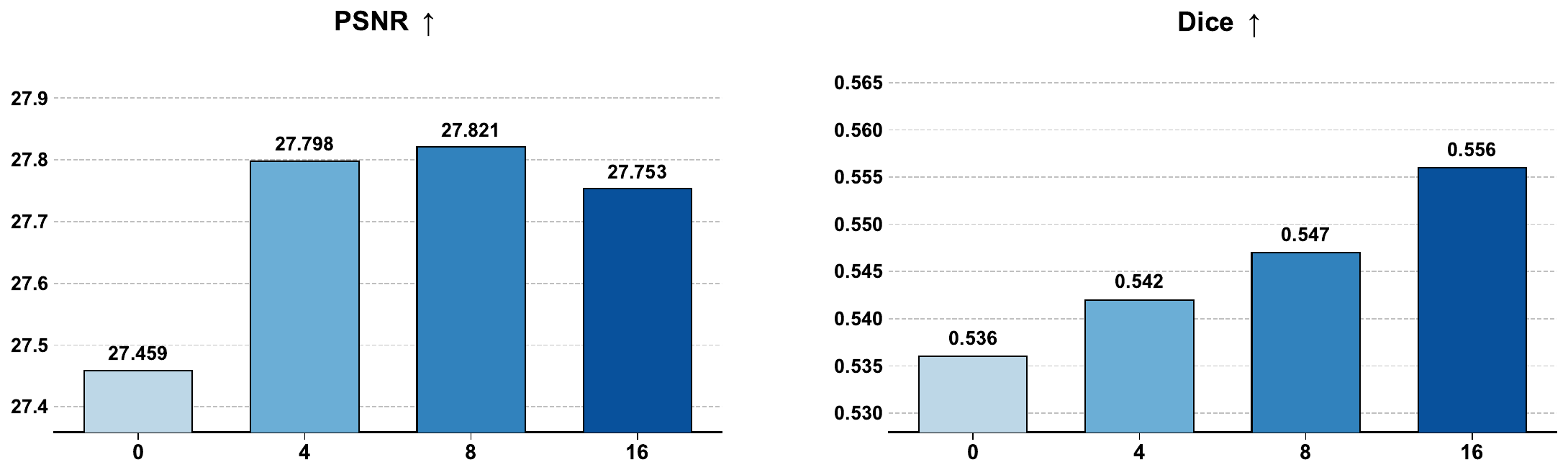}
    \caption{
    Detailed ablation of the number of organ-specific memories $L$ in \modelname{}. Models are trained without \alignname{} for 4k iterations.
    }
    \label{fig:suppl_ablation_detail}
    % \vspace{-8pt}
\end{figure}
Fig.~\ref{fig:suppl_ablation_detail} presents the ablation on the number of organ-specific memories $L$ in \modelname{}. As $L$ increases, the Dice score improves consistently, while the PSNR peaks at $L=8$ and slightly decreases at $L=16$. To balance image fidelity and enhancement accuracy, we set $L=8$ in our model.

%% file: secs/discuss_conclu.tex
\begin{figure}[t]
    \centering    
    \includegraphics[width=1.0\linewidth]{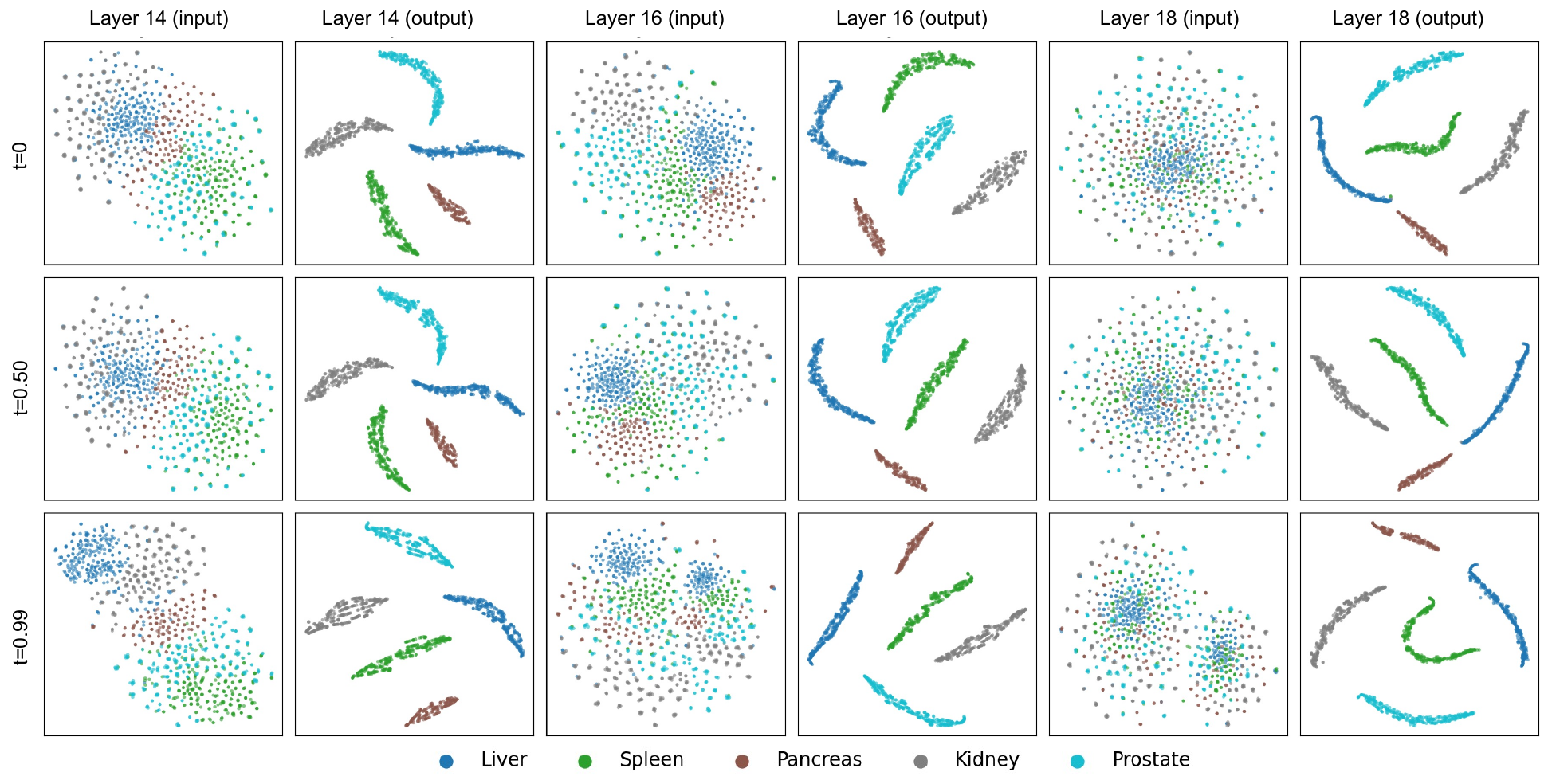}
    \caption{t-SNE visualization of layer-wise input and output representations of \modelname{} at different timesteps $ t $ during inference stage. }
    \label{fig:tsne} 
    % \vspace{-8pt}
\end{figure}

\paragraph{t-SNE visualization.} We visualize the input and output features of \modelname{} at various timesteps using t-SNE, as shown in Fig.~\ref{fig:tsne}. Since the NCCT image serves as a condition input, we observe that the input features exhibit relatively clear organ-wise clustering in earlier layers (\eg, Layer 14), while becoming increasingly mixed in deeper layers (\eg, Layer 18) due to the progressive feature abstraction. In contrast, after passing through each expert, the outputs consistently exhibit well-separated clustering by organ type across all layers and timesteps, demonstrating their effectiveness in maintaining organ-specific representations throughout the network.

\section{Discussion}
% \section{Discussion and Conclusion}
\label{sec:discussion}

This paper introduces \methodname, a volumetric diffusion framework for high-fidelity virtual contrast enhancement.

First, to our knowledge, we are the first to leverage video diffusion models for CT VCE. Compared to other medical imaging modalities such as MRI and ultrasound, which typically have resolutions around 200$ \times $200, CT images usually have a resolution of 512$ \times $512. By treating multiple CT slices as video frames, \methodname effectively handles CT's high-resolution demands while improving inter-slice coherence.

Second, unlike natural images where out-of-distribution object categories can be virtually unlimited, human anatomy consists of a finite set of organs. Therefore, \methodname proposes organ-level experts rather than requiring the model to implicitly learn anatomical variations. By doing so, we observe significant improvements in enhancement accuracy. Furthermore, \methodname is extensible---future work could incorporate disease-specific experts with targeted segmentation, \eg, vascular structures.

Third, we find that contrast-enhanced features are subtle intensity variations superimposed on anatomical structures. Naive image-to-image translation tends to learn anatomical mappings rather than fine-grained contrast features. To address this, we propose \alignname to reinforce contrast-aware representations, which notably improves low-level metrics.
Overall, \methodname achieves state-of-the-art performance across multiple datasets, balancing image fidelity and enhancement accuracy with strong generalization capability, providing practical guidance for future CT VCE research.

% \paragraph{Limitations.}
We also acknowledge some limitations. First, \methodname relies on segmentation masks at inference, introducing an additional preprocessing step, though strong off-the-shelf models are available~\cite{totalsegmentator,medsam2}. Second, while \alignname mitigates imperfect alignment, residual misalignment can still cause synthesis errors, especially for hollow organs with large non-rigid deformations. Third, training video diffusion models is computationally demanding; improving efficiency through model sparsification and faster solvers remains desirable for broader deployment.

\section{Conclusion}
In conclusion, this paper introduces \methodname{}, a volumetric diffusion framework for high-fidelity virtual contrast enhancement. By treating CT volumes as coherent sequences and introducing \modelname{} and \alignname{} to address anatomical heterogeneity and spatial misalignment, \methodname{} significantly enhance synthesis quality and enhancement accuracy.